%% file: arXiv.tex
\documentclass{article} 
\usepackage{iclr2027_conference,times}

\usepackage{graphicx}
\usepackage{times}
\usepackage{helvet}
\usepackage{courier}
\usepackage{amsmath}
\usepackage{algorithm}
\usepackage{algorithmic}
\usepackage{csquotes} 
\usepackage{color}
\usepackage{paralist}
\usepackage{amssymb}
\usepackage{indentfirst}
\usepackage{float}
\usepackage{multirow}
\usepackage{cite}
\usepackage{mathrsfs}

\usepackage{mathrsfs}
\usepackage{textcomp,booktabs}
\usepackage{amssymb}
\usepackage{pifont}
\usepackage[misc]{ifsym}
\usepackage{caption}
\usepackage{mathrsfs}
\usepackage{color}
\usepackage{xcolor}

\input{math_commands.tex}

\usepackage[hidelinks]{hyperref}
\usepackage{url}

\title{UniMergeTrack: Hierarchical Model Merging for General RGB-X Visual Object Tracking}

\author{Shiao Wang$^{1}$, Xiao Wang$^{1}$~\thanks{Corresponding author: Xiao Wang}, 
Yuhe Ding$^{1}$, Jiarui Xu$^{1}$, Zikang Yan$^{1}$, Bo Jiang$^{1}$, Yonghong Tian$^{2,3,4}$, Jin Tang$^{1}$ \\ 
1. School of Computer Science and Technology, Anhui University, Hefei 230601, China \\ 
2. Peng Cheng Laboratory, Shenzhen, China \\ 
3. School of Computer Science, Peking University, China \\ 
4. School of Electronic and Computer Engineering, Shenzhen Graduate School, Peking University, China \\ 
\texttt{wsa1943230570@126.com, \{xiaowang, tangjin, jiangbo\}@ahu.edu.cn} \\ 
\texttt{17809185626@163.com, madao3c@foxmail.com} \\ 
\texttt{e24201056@stu.ahu.edu.cn, yhtian@pku.edu.cn} 
}

\iclrfinalcopy 
\begin{document}

\maketitle

\begin{abstract}
RGB-X tracking leverages complementary modalities, such as thermal infrared, event, or depth, to improve robustness under challenging conditions. Existing general-purpose RGB-X trackers typically follow two paradigms: independent task-specific training and joint training across modality combinations. However, the former suffers from knowledge isolation and model redundancy, while the latter introduces optimization interference across heterogeneous tasks. In this work, we present UniMergeTrack, a new model merging paradigm that preserves task-specific optimization while enabling effective cross-task knowledge integration. UniMergeTrack first trains modality-specific trackers independently and then performs hierarchical model merging. Specifically, transferable backbones are merged across RGB-X tasks via high-pass-filtered merging, and complemented by lightweight task-specific experts that preserve modality-specific cues. Tracking heads are merged within each task using optimal transport-guided channel alignment to mitigate hidden channel mismatch. A subsequent distribution alignment fine-tuning stage further improves compatibility between the unified backbone and task-specific heads. Extensive experiments on the LasHeR, RGBT234, VisEvent, VOT-RGBD22, COESOT, and CDTB benchmarks demonstrate that UniMergeTrack achieves state-of-the-art or competitive performance across RGB-Thermal, RGB-Event, and RGB-Depth tracking tasks, while exhibiting strong generalization to unseen datasets. The source code will be released upon acceptance.
The source code of this paper will be released on \url{https://github.com/Event-AHU/OpenEvTracking}.
\end{abstract}

\section{Introduction}

Visual object tracking (VOT) aims to continuously localize a target in a video given its initial state, serving as a fundamental component in various vision applications, including autonomous driving, surveillance, and human-computer interaction.
Deep learning has driven significant advances in RGB-based trackers~\citep{yan2021stark,cui2022mixformer,ye2022joint,chen2023seqtrack}. However, RGB cameras remain vulnerable to adverse conditions such as low illumination, motion blur, occlusion, and background clutter, where appearance cues may become ambiguous or unreliable. Auxiliary modalities, including thermal infrared, event, and depth, provide complementary information beyond visible appearance. Multimodal tracking has therefore gained increasing attention, with early methods tailored to specific modality combinations such as RGB-Thermal~\citep{wang2020cmpp,zhang2021jmmac}, RGB-Event~\citep{zhang2021fenet,wang2024visevent}, and RGB-Depth~\citep{kart2019otr,yan2021depthtrack}, but generalizing poorly across different modality combinations. 

\begin{figure*}[!htp]
\center
\includegraphics[width=\linewidth]{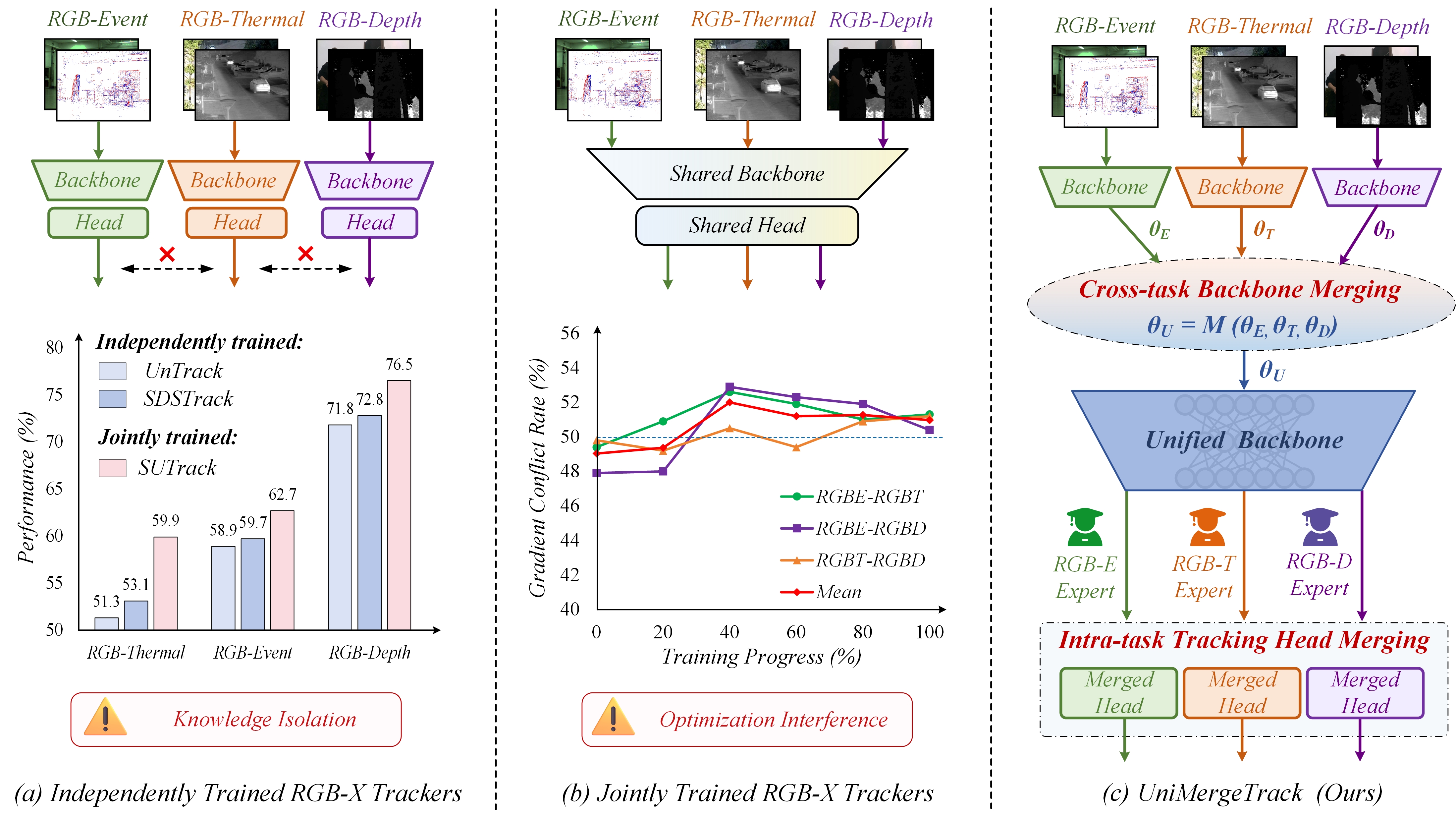}
\caption{\textbf{Comparison of three paradigms for RGB-X tracking.} (a) Independent training retains task specialization but suffers from knowledge isolation. (b) Joint training enables knowledge sharing but introduces optimization interference. (c) UniMergeTrack achieves unified tracking via hierarchical model merging with cross-task backbone and intra-task head merging.} 
\label{First_Image}
\end{figure*}

More recently, general RGB-X (X denotes thermal infrared, event, or depth data) multimodal trackers have been developed to handle multiple modality combinations within a unified framework. Existing approaches mainly follow two paradigms. The first adopts a general tracking architecture but trains it independently for each modality combination~\citep{zhu2023vipt,hou2024sdstrack,wu2024untrack}, resulting in multiple task-specific trackers. As shown in Fig.~\ref{First_Image} (a), although such independent training preserves task-specific specialization, it leads to model redundancy and knowledge isolation across RGB-X tasks. The second jointly trains a single model on mixed RGB-X data~\citep{chen2025sutrack,hong2026onetrackerv2,geng2026tgtrack}, enabling shared representations and improved cross-task knowledge transfer. However, as illustrated in Fig.~\ref{First_Image} (b), joint training suffers from persistent gradient conflicts across tasks (measured by the Gradient Conflict Rate defined based on negative gradient cosine similarity~\citep{yu2020pcgrad,gong2022nasvit}), suggesting optimization interference among heterogeneous RGB-X tasks and potentially constraining task-specific specialization.

These complementary limitations motivate us to explore an alternative paradigm that preserves task specialization while enabling effective cross-task knowledge integration. To this end, we introduce \textbf{UniMergeTrack}, a model merging paradigm for general RGB-X tracking, as illustrated in Fig.~\ref{First_Image} (c). Instead of jointly optimizing heterogeneous modality combinations in a shared parameter space, UniMergeTrack first trains modality-specific trackers independently and then integrates their knowledge through hierarchical model merging. Specifically, we merge backbone parameters across RGB-X tasks via high-pass-filtered merging to construct a unified backbone, complemented by lightweight task-specific experts that retain modality-specific cues. In contrast, tracking heads are more closely coupled to task-specific prediction behavior; therefore, we merge them within each task using optimal transport-guided channel alignment. Finally, distribution alignment fine-tuning is performed to mitigate the distribution shifts introduced by model merging and ensure compatibility between the unified backbone and task-specific tracking heads.
In this way, UniMergeTrack retains task-specific specialization while integrating cross-task knowledge, providing an effective alternative to conventional paradigms.

To sum up, the contributions of this work can be summarized as follows:
1) We present UniMergeTrack, the first model merging paradigm for RGB-X tracking, which alleviates knowledge isolation and optimization interference in conventional paradigms.
2) We propose a hierarchical model merging strategy, with high-pass-filtered cross-task merging for backbones and OT-guided intra-task merging for tracking heads.
3) Extensive experiments on multiple RGB-X tracking benchmarks, including LasHeR, RGBT234, VisEvent, VOT-RGBD22, COESOT, and CDTB, demonstrate the effectiveness and generalization ability of the proposed tracker.

\section{Related Work} 

\subsection{RGB-X Tracking}
RGB-X tracking leverages an auxiliary modality, such as thermal infrared, event, or depth data, to complement RGB observations under challenging conditions. Early multimodal trackers were typically tailored to a specific modality combination, improving robustness within their target settings. Representative methods include dedicated RGB-Event trackers~\citep{zhang2021fenet,wang2024visevent,tang2025revisiting}, RGB-Thermal trackers~\citep{wang2020cmpp,zhang2021jmmac,xiao2022apfnet}, and RGB-Depth trackers~\citep{kart2019otr,qian2020dal,yan2021depthtrack}. However, such modality-specific designs generalize poorly across modality combinations.
Recent works have therefore explored general RGB-X tracking. Existing approaches mainly follow two paradigms. The first employs a shared tracking framework while training it separately for each modality combination, as in ViPT~\citep{zhu2023vipt}, SDSTrack~\citep{hou2024sdstrack}, and OneTracker~\citep{hong2024onetracker}. The second jointly trains a single model on mixed data from multiple modality combinations, enabling one parameter set to handle diverse RGB-X tracking tasks, as in SUTrack~\citep{chen2025sutrack}, OneTrackerV2~\citep{hong2026onetrackerv2}, and TGTrack~\citep{geng2026tgtrack}.
While both paradigms improve generalization across modality combinations, they exhibit complementary limitations. Independent adaptation leads to model redundancy and limited knowledge sharing across tasks, whereas joint training suffers from optimization interference among heterogeneous modality combinations, potentially constraining task-specific specialization. In contrast, we introduce a new model merging paradigm for RGB-X tracking, which combines the strengths of both approaches while mitigating knowledge isolation and optimization interference.

\subsection{Model Merging}
Model merging consolidates multiple fine-tuned models into a single model, reducing redundancy while improving cross-task generalization. Early studies established weight-space averaging as a simple yet effective strategy. Stochastic Weight Averaging (SWA)~\citep{izmailov2018swa} averages checkpoints along an optimization trajectory, while Model Soups~\citep{wortsman2022modelsoups} extends this idea to models fine-tuned from a shared pretrained initialization.
Recent methods increasingly focus on merging task-specific models derived from a common pretrained model. Fisher Merging~\citep{matena2022fisher} weights parameters by their estimated importance. Task Arithmetic~\citep{ilharco2023editing} represents task knowledge as parameter displacements from the shared initialization and composes them through vector arithmetic. To mitigate redundancy and destructive interference, TIES-Merging~\citep{yadav2023ties} trims insignificant updates and resolves sign conflicts, while DARE~\citep{yu2024dare} sparsifies task vectors through random dropping and rescaling. FREE-Merging~\citep{zheng2025free} further addresses task interference in the frequency domain, whereas RAIN-Merging~\citep{huang2026rain} employs null-space projection and module-specific scaling to preserve existing capabilities.
Despite these advances, most methods remain architecture-agnostic, applying the same merging rule to transferable backbones and task-specific heads, potentially causing information misalignment. We adopt a hierarchical merging strategy that merges backbones across tasks and tracking heads within each task.

\begin{figure*}[!htp]
\center
\includegraphics[width=\linewidth]{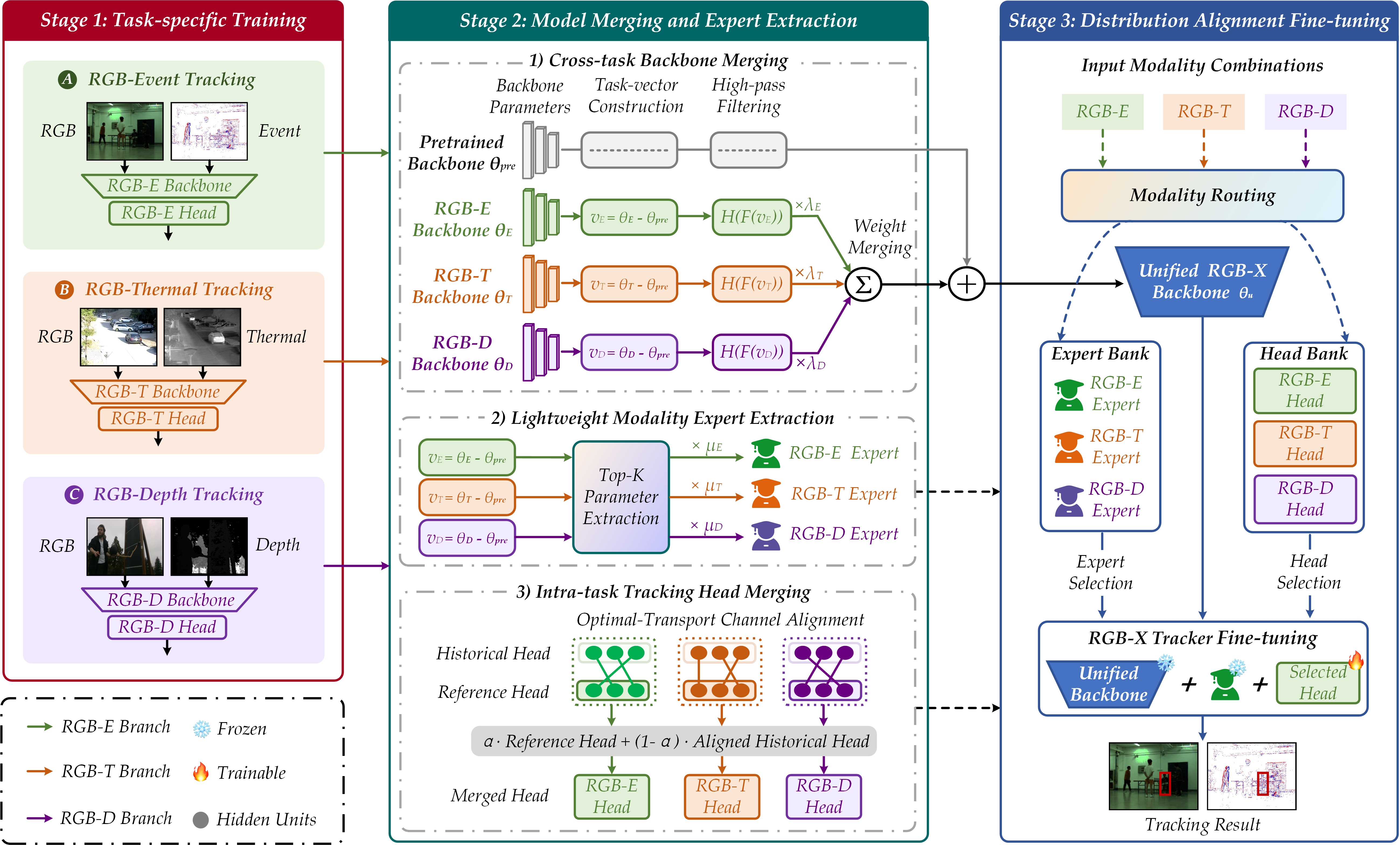}
\caption{\textbf{Overview of the proposed model merging framework for RGB-X tracking.} We first train RGB-E, RGB-T, and RGB-D trackers independently, then perform hierarchical model merging by merging high-pass-filtered backbone task vectors across tasks and OT-aligned tracking heads within each task, while extracting lightweight modality experts. Finally, distribution alignment fine-tuning improves the compatibility between the unified backbone and task-specific tracking heads.}
\label{framework}
\end{figure*}

\section{Our Proposed Approach}

\subsection{Preliminary}
We build on Task Arithmetic (TA)~\citep{ilharco2023editing}, which represents task-specific knowledge as parameter shifts from a shared pretrained model. Let \(\boldsymbol{\theta}_{pre}\) and \(\boldsymbol{\theta}_k\) denote the pretrained and task-\(k\) fine-tuned parameters, respectively. The task vector is defined as
\begin{equation}
\boldsymbol{v}_k
=
\boldsymbol{\theta}_k
-
\boldsymbol{\theta}_{pre}.
\end{equation}
Given $K$ task-specific models, their acquired knowledge can be integrated by performing arithmetic operations on the corresponding task vectors:
\begin{equation}
\boldsymbol{\theta}_{\mathrm{merge}}
=
\boldsymbol{\theta}_{pre}
+
\sum_{k=1}^{K}
\lambda_k \boldsymbol{v}_k,
\end{equation}
where $\lambda_k$ controls the contribution of the $k$-th task vector. 
This formulation enables knowledge acquired from multiple tasks to be composed without joint training on their combined datasets. 

\subsection{Overview}
We propose UniMergeTrack, a model merging framework for general RGB-X tracking. In the first stage, we train RGB-Event, RGB-Thermal, and RGB-Depth trackers independently. In the second stage, we merge backbone parameters across tasks while extracting lightweight task-specific experts, and merge tracking heads within each task. In the third stage, we freeze the merged backbone and experts, fine-tuning tracking heads and other lightweight parameters to improve the compatibility between the unified backbone and tracking heads. The overall framework is illustrated in Fig.~\ref{framework}.

\subsection{Task-specific Training}
To support multiple multimodal SOT tasks within a unified one-stream framework, we initialize all task-specific models from the same pretrained weights and independently train them on the corresponding RGB-Event, RGB-Thermal, and RGB-Depth datasets.

Following SUTrack~\citep{chen2025sutrack}, we concatenate an input RGB image $\mathbf{I}_{\mathrm{RGB}} \in \mathbb{R}^{H \times W \times 3}$ with its corresponding X-modality image $\mathbf{I}_{\mathrm{X}} \in \mathbb{R}^{H \times W \times 3}$ along the channel dimension, producing a multimodal input $\mathbf{I}_{\mathrm{concat}} \in \mathbb{R}^{H \times W \times 6}$, where X represents an event, thermal infrared, or depth modality. The concatenated input is then partitioned into non-overlapping patches of size $P \times P$, each of which is flattened into a $6P^2$-dimensional vector. Let $\mathbf{F}_{p} \in \mathbb{R}^{N \times 6P^2}$ denote the resulting sequence of flattened patch vectors, where $N$ is the number of patches. These vectors are mapped into a $D$-dimensional embedding space through a learnable linear projection:
\begin{equation}
\mathbf{E}
=
\mathbf{F}_{p}\mathbf{W}_{\mathrm{proj}}
+
\mathbf{b}_{\mathrm{proj}},
\end{equation}
where $\mathbf{W}_{\mathrm{proj}} \in \mathbb{R}^{6P^2 \times D}$ and $\mathbf{b}_{\mathrm{proj}} \in \mathbb{R}^{D}$ denote the projection matrix and bias vector, respectively. This projection produces the multimodal patch embeddings $\mathbf{E} \in \mathbb{R}^{N \times D}$.

The template and search regions are embedded identically, concatenated along the token dimension, and fed into the Transformer backbone for joint feature extraction and cross-modal interaction. The resulting features are passed to a center-based tracking head for target prediction. We use this pipeline to independently train RGB-Event, RGB-Thermal, and RGB-Depth trackers.

\subsection{Model Merging and Expert Extraction} 

\textbf{Cross-task Backbone Merging.~}
Given the independently trained backbone parameters $\boldsymbol{\theta}_{E}$, $\boldsymbol{\theta}_{T}$, and $\boldsymbol{\theta}_{D}$ for RGB-Event, RGB-Thermal, and RGB-Depth tracking, respectively, we integrate their knowledge into a unified backbone that supports all three modality combinations. Since these backbones share the same pretrained initialization $\boldsymbol{\theta}_{pre}$, we construct the corresponding task vectors as:
\begin{equation}
    \boldsymbol{v}_{k}
    =
    \boldsymbol{\theta}_{k}
    -
    \boldsymbol{\theta}_{pre},
    \qquad
    k\in \{E, T, D\},
\end{equation}
where $\boldsymbol{v}_{k}$ characterizes the parameter displacement induced by task-specific training.

Directly merging task vectors can induce cross-task interference~\citep{yadav2023ties} due to conflicting updates across different modality combinations. As low-frequency components are particularly susceptible to such interference~\citep{zheng2025free}, we apply high-pass filtering to each task vector before merging to suppress these conflict-prone components.

For the $l$-th parameter tensor, we first transform each task-vector slice into the frequency domain:
\begin{equation}
    \widehat{\boldsymbol{v}}_{k}^{\,l}
    =
    \mathcal{F}\!\left(
        \boldsymbol{v}_{k}^{\,l}
    \right),
    \label{eq:fourier_task_vector}
\end{equation}
where $\mathcal{F}(\cdot)$ denotes the discrete Fourier transform. We then apply a radial high-pass mask $\mathbf{H}_{k}^{\,l}$:
\begin{equation}
\mathbf{H}_{k}^{\,l}(u,v)
=
\begin{cases}
    1, & A_l(u,v) \geq \rho_l,\\
    0, & A_l(u,v) < \rho_l,
\end{cases}
\qquad
\rho_l = \eta \cdot \min(H_l,W_l).
\label{eq:high_pass_mask}
\end{equation}
where $A_l(u,v)$ is the radial distance from frequency coordinate $(u,v)$ to the spectral center, $\rho_l$ is the cutoff radius, and $H_l$ and $W_l$ are the spatial dimensions of the frequency representation. We set the filtering ratio $\eta$ to $0.7$ by default. The filtered task vector is then obtained as
\begin{equation}
    \widetilde{\boldsymbol{v}}_{k}^{\,l}
    = \mathcal{F}^{-1}\!\left(
            \mathbf{H}_{k}^{\,l}
            \odot
            \widehat{\boldsymbol{v}}_{k}^{\,l}
        \right),
    \label{eq:filtered_task_vector}
\end{equation}
where $\odot$ denotes element-wise multiplication and $\mathcal{F}^{-1}(\cdot)$ is the inverse Fourier transform. This filtering suppresses low-frequency interference while preserving high-frequency variations.

Finally, the unified RGB-X backbone is constructed by merging the filtered task vectors into the shared pretrained backbone:
\begin{equation}
\boldsymbol{\theta}_{U}
=
\boldsymbol{\theta}_{pre}
+
\sum_{k \in \{E,T,D\}}
\lambda_k \widetilde{\boldsymbol{v}}_{k},
\end{equation}
where $\lambda_k$ denotes the merging coefficient, which is set to 0.25 by default.

\textbf{Lightweight Modality Expert Extraction.~}
Given a task vector $\boldsymbol{v}_{k}$, where $k \in \{E,T,D\}$, the magnitude of fine-tuning-induced parameter updates provides a natural measure of task specialization, with larger updates typically reflecting stronger task-specific adaptation. To extract a sparse modality expert, $R_{\gamma}(\cdot)$ first removes the top $\gamma$ fraction of entries by magnitude, and $M_{d}(\cdot)$ then retains the top $d$ fraction of the remaining entries. The sparse task expert is defined as
\begin{equation}
    \hat{\boldsymbol{E}}_{k}
    =
    M_{d}\left(R_{\gamma}\left(\boldsymbol{v}_{k}\right)\right),
\end{equation}
where $d \ll 1$ controls the sparsity of the expert. When $\gamma = 0$, this formulation reduces to standard magnitude-based Top-$d$ selection. 

Although sparse extraction reduces parameter overhead, it alters the statistics of the original task vector and may weaken the expert contribution when combined with the unified backbone. To compensate, we further introduce a task-wise rescaling strategy. Let $\bar{\boldsymbol{v}}_{k}$ denote the shared task vector used in backbone merging, and $\mathcal{E}(\cdot)$ denote the mean absolute parameter value averaged across parameter tensors. The rescaling factor and the final task expert are defined as
\begin{equation}
    \mu_{k}
    =
    -\frac{
        \mathcal{E}\!\left(\hat{\boldsymbol{E}}_{k}\right)\log d
    }{
        d\,\mathcal{E}\!\left(\bar{\boldsymbol{v}}_{k}\right)
    },
    \qquad
    \boldsymbol{E}_{k}
    =
    \mu_{k}\hat{\boldsymbol{E}}_{k}.
\end{equation}

This rescaling compensates for the magnitude shift caused by sparse extraction, preserving the contribution of retained updates under high sparsity. Thus, each RGB-X task requires only a lightweight expert rather than a separate backbone. The resulting expert bank,
$\mathcal{B}_E =
    \left\{
        \boldsymbol{E}_{E},
        \boldsymbol{E}_{T},
        \boldsymbol{E}_{D}
    \right\},$
enables task-specific adaptation to the unified backbone with minimal storage overhead.

\textbf{Intra-task Tracking Head Merging.~}
Unlike the backbone, which learns transferable representations and benefits from cross-task merging, the tracking head is closely tied to task-specific prediction distributions. Cross-task head merging may therefore introduce interference across heterogeneous objectives. We instead retain a separate head for each task and merge checkpoints within the same task to consolidate knowledge from different optimization stages.

However, even heads trained for the same task may develop inconsistent hidden-channel organizations, making direct parameter merging unreliable. For each task $k$, we align a historical head $\boldsymbol{h}_{k}^{\mathrm{his}}$ with the final reference head $\boldsymbol{h}_{k}^{\mathrm{ref}}$ using optimal transport (OT)~\citep{cuturi2013sinkhorn}. Each output channel is represented by a neuron descriptor constructed from normalized incoming weights, and the pairwise channel cost is defined as
\begin{equation}
    \mathbf{C}_{ij}
    =
    \left\|
    \boldsymbol{d}_{i}^{\mathrm{his}}
    -
    \boldsymbol{d}_{j}^{\mathrm{ref}}
    \right\|_{2}^{2},
\end{equation}
where $\boldsymbol{d}_{i}^{\mathrm{his}}$ and $\boldsymbol{d}_{j}^{\mathrm{ref}}$ denote the descriptors of the $i$-th historical and $j$-th reference channels, respectively. We then solve the entropy-regularized OT problem
\begin{equation}
    \mathbf{P}^{*}
    =
    \arg\min_{\mathbf{P}\geq0}
    \langle \mathbf{P},\mathbf{C}\rangle_F
    -
    \epsilon H(\mathbf{P}),
    \quad
    \mathrm{s.t.}\ 
    \mathbf{P}\mathbf{1}
    =
    \frac{1}{B_s}\mathbf{1},
    \quad
    \mathbf{P}^{\top}\mathbf{1}
    =
    \frac{1}{B_t}\mathbf{1},
\end{equation}
where $\mathbf{P}^{*}\in\mathbb{R}^{B_s\times B_t}$ is the optimal transport plan, $B_s$ and $B_t$ are the numbers of historical and reference channels, and
$H(\mathbf{P})=-\sum_{i,j}P_{ij}\log P_{ij}$ is the entropy term with regularization coefficient $\epsilon>0$. We obtain $\mathbf{P}^{*}$ via Sinkhorn iterations~\citep{cuturi2013sinkhorn} and propagate the resulting correspondence layer by layer to align the historical head with the reference head before merging.
After that, the aligned historical head is merged with the reference head through weighted parameter averaging:
\begin{equation}
    \boldsymbol{H}_{k}
    =
    \alpha \boldsymbol{h}_{k}^{\mathrm{ref}}
    +
    (1-\alpha)\widetilde{\boldsymbol{h}}_{k}^{\mathrm{his}},
    \label{eq:head_merging}
\end{equation}
where $\widetilde{\boldsymbol{h}}_{k}^{\mathrm{his}}$ denotes the OT-aligned historical head and $\alpha$ controls the contribution of the reference head. The resulting task-specific heads form the head bank
$    \mathcal{B}_{H}
    =
    \left\{
        \boldsymbol{H}_{E},
        \boldsymbol{H}_{T},
        \boldsymbol{H}_{D}
    \right\}.$

\subsection{Distribution Alignment Fine-tuning}
Although model merging integrates shared and modality-specific knowledge, the recomposed components may not be fully compatible. In particular, replacing the original task-specific backbone with the unified backbone shifts the feature distribution seen by the tracking head. We therefore perform lightweight post-merging fine-tuning to improve their compatibility.

For each modality $k \in \{E,T,D\}$, we compose the corresponding lightweight expert $\boldsymbol{E}_{k}$ with the unified backbone $\boldsymbol{\theta}_U$ through sparse parameter overriding, yielding a task-specialized backbone:
\begin{equation}
    \left(\boldsymbol{\theta}^{k}_{U}\right)_i
    =
    \begin{cases}
        \left(\boldsymbol{E}_{k}\right)_i, & i \in \mathcal{I}_{k}, \\
        \left(\boldsymbol{\theta}_{U}\right)_i, & i \notin \mathcal{I}_{k},
    \end{cases}
    \label{eq:specialized_backbone}
\end{equation}
where $\mathcal{I}_{k}$ denotes the sparse parameter indices retained by the task expert.
The resulting backbone integrates the cross-task knowledge retained by the unified backbone with the complementary task-specific knowledge encoded by the lightweight experts. 

We pair each backbone $\boldsymbol{\theta}^{k}_{U}$ with its corresponding tracking head $\boldsymbol{H}_{k}$, and independently fine-tune the resulting tracker on the task-specific training distribution $\mathcal{D}_k$. The optimization objective is
\begin{equation}
    \min_{\boldsymbol{\phi}_k}
    \mathbb{E}_{(\mathbf{x},\mathbf{y})\sim\mathcal{D}_k}
    \left[
        \mathcal{L}_{\mathrm{total}}
        \left(
            f_k(\mathbf{x};
            \boldsymbol{\theta}^{k}_{U},
            \boldsymbol{\phi}_k),
            \mathbf{y}
        \right)
    \right],
    \qquad
    \boldsymbol{\theta}^{k}_{U}\ \text{frozen},
    \quad
    k \in \{E,T,D\},
    \label{eq:distribution_alignment}
\end{equation}
where \(\boldsymbol{\phi}_k\) comprises the trainable tracking head and other lightweight task-specific parameters, while the unified backbone remains frozen to prevent its consolidated parameters from being overwritten. This stage enhances the compatibility between the backbone and its task-specific head.

\subsection{Tracking Head and Loss Function}
Following SUTrack~\citep{chen2025sutrack}, we adopt a center-based tracking head to predict the target location and bounding box from the search features. The tracking objective consists of a weighted focal loss $\mathcal{L}_{\mathrm{cls}}$ for target classification, together with the generalized IoU loss $\mathcal{L}_{\mathrm{GIoU}}$ and $\ell_1$ loss $\mathcal{L}_{\ell_1}$ for bounding-box regression. We further employ a cross-entropy loss $\mathcal{L}_{\mathrm{task}}$ for task recognition. The overall objective is
\begin{equation}
    \mathcal{L}_{\mathrm{total}}
    =
    \mathcal{L}_{\mathrm{cls}}
    +
    \lambda_{\mathrm{GIoU}}\mathcal{L}_{\mathrm{GIoU}}
    +
    \lambda_{\ell_1}\mathcal{L}_{\ell_1}
    +
    \lambda_{\mathrm{task}}\mathcal{L}_{\mathrm{task}},
\end{equation}
where $\lambda_{\mathrm{GIoU}}=2$, $\lambda_{\ell_1}=5$, and $\lambda_{\mathrm{task}}=1$. The same objective and loss weights are used for both initial training and post-merging fine-tuning.

\section{Experiments} 


\subsection{Implementation Details} 
\noindent \textbf{Model Architecture.}
UniMergeTrack is built upon the SUTrack~\citep{chen2025sutrack} architecture and adopts HiViT-base~\citep{zhang2023hivit} for patch embedding and backbone feature extraction. 
The model is initialized with weights~\citep{chen2026relo} pretrained on LaSOT~\citep{fan2019lasot}, TrackingNet~\citep{muller2018trackingnet}, COCO~\citep{lin2014coco}, VastTrack~\citep{peng2024vasttrack}, and GOT-10k~\citep{huang2021got10k}.
The model is implemented in Python 3.8 with PyTorch 1.11. All experiments are conducted on a single NVIDIA RTX 4090 GPU.

\noindent \textbf{Training Setup.} 
\textit{For the task-specific training stage}, we independently train RGB-Thermal, RGB-Event, and RGB-Depth models on LasHeR~\citep{li2021lasher}, VisEvent~\citep{wang2024visevent}, and DepthTrack~\citep{yan2021depthtrack}, respectively. To ensure a fair comparison with existing RGB-X trackers, we additionally include RGB data from TNL2K~\citep{wang2021tnl2k} during this stage. The template and search regions are resized to $112\times112$ and $224\times224$, respectively. Each task is trained for 30 epochs, with 100,000 training pairs sampled per epoch. We adopt AdamW~\citep{loshchilov2019decoupled} as the optimizer, with initial learning rates of $1\times10^{-5}$ for the Transformer encoder and $1\times10^{-4}$ for the remaining parameters. The learning rates are decayed by a factor of 10 after 24 epochs, with a weight decay of $1\times10^{-4}$.
\textit{During the distribution alignment fine-tuning stage}, the unified backbone is frozen, while the tracking heads are separately fine-tuned on their corresponding RGB-X tasks. TNL2K is excluded at this stage, and all other training settings remain unchanged.

\subsection{Comparison on Public Benchmark Datasets} 

$\bullet$ \textbf{Results on LasHeR Dataset.}
As shown in Table~\ref{tab:rgbx_comparison}, we compare UniMergeTrack with existing mainstream trackers on the LasHeR dataset. UniMergeTrack achieves an AUC of 61.8\% and a PR of 77.2\%, attaining the best AUC among all compared methods. Compared with the independently trained XTrack and the jointly trained SUTrack, our method improves PR by 8.1 and 2.7 percentage points, respectively. These results demonstrate that our model merging paradigm effectively integrates cross-task knowledge while preserving task-specific tracking capability.

$\bullet$ \textbf{Results on RGBT234 Dataset.}
On the RGBT234 dataset, UniMergeTrack achieves 70.6\% MSR and 94.0\% MPR, ranking first on both metrics. Compared with XTrack, it improves MSR and MPR by 5.7 and 6.6 percentage points, respectively, and outperforms SUTrack by 1.1 and 1.8 points. These consistent gains further validate the effectiveness of our method for RGB-Thermal tracking.

$\bullet$ \textbf{Results on VisEvent Dataset.}
As illustrated in Table~\ref{tab:rgbx_comparison}, we also report results on the RGB-Event tracking task. UniMergeTrack achieves 64.1\% AUC and 81.5\% PR on VisEvent, establishing new state-of-the-art performance on both metrics. It surpasses the jointly trained SUTrack by 1.6 percentage points in PR and the recent OneTrackerV2 by 1.7 percentage points. These results further demonstrate the effectiveness of UniMergeTrack in exploiting motion cues from event data.

$\bullet$ \textbf{Results on VOT-RGBD22 Dataset.}
We further evaluate UniMergeTrack on VOT-RGBD22 for RGB-Depth tracking. As shown in Table~\ref{tab:rgbx_comparison}, it achieves 76.8\% EAO, 83.1\% Accuracy, and 92.3\% Robustness, obtaining the best Accuracy and competitive overall performance. These results further validate the effectiveness of UniMergeTrack for RGB-Depth tracking.

$\bullet$ \textbf{Generalization Performance on COESOT~\citep{tang2025revisiting} and CDTB~\citep{Lukezic_2019_ICCV}.~}
To assess generalization, we directly evaluate UniMergeTrack and recent open-source trackers on COESOT and CDTB without any additional training, as shown in Table~\ref{tab:generalization}. UniMergeTrack achieves 66.3\% SR and 80.5\% PR on COESOT, outperforming SUTrack by 1.7 and 2.8 percentage points, respectively. On CDTB, UniMergeTrack also achieves the best performance in terms of both F-score and Precision, further demonstrating its robust generalization ability to unseen datasets.

\begin{table*}[t]
\centering
\caption{Comparison results on LasHeR, RGBT234, VisEvent, and VOT-RGBD22 datasets. The best and second-best results are highlighted in \textbf{bold} and \underline{underlined}, respectively.}
\label{tab:rgbx_comparison}
\renewcommand{\arraystretch}{1.05}
\setlength{\tabcolsep}{7pt}
\resizebox{\textwidth}{!}{
\begin{tabular}{l|c|cc|cc|cc|ccc}
\toprule
\multirow{2}{*}{Method} & \multirow{2}{*}{Source} & \multicolumn{2}{c|}{LasHeR} & \multicolumn{2}{c|}{RGBT234} & \multicolumn{2}{c|}{VisEvent} & \multicolumn{3}{c}{VOT-RGBD22} \\
\cline{3-11}
& & AUC & PR & MSR & MPR & AUC & PR & EAO & Acc. & Rob. \\
\midrule
OSTrack~\citep{ye2022joint} & ECCV22 & 41.2 & 51.5 & 54.9 & 72.9 & 53.4 & 69.5 & 67.6 & 80.3 & 83.3 \\
ProTrack~\citep{yang2022protrack} & ACM MM22 & 42.0 & 53.8 & 59.9 & 79.5 & 47.1 & 63.2 & 65.1 & 80.1 & 80.2 \\
ViPT~\citep{zhu2023vipt} & CVPR23 & 52.5 & 65.1 & 61.7 & 83.5 & 59.2 & 75.8 & 72.1 & 81.5 & 87.1 \\
SeqTrackv2~\citep{chen2023seqtrackv2} & ArXiv23 & 55.8 & 70.4 & 64.7 & 88.0 & 61.2 & 78.2 & 74.4 & 81.5 & 91.0 \\
OneTracker~\citep{hong2024onetracker} & CVPR24 & 53.8 & 67.2 & 64.2 & 85.7 & 60.8 & 76.7 & 72.7 & 81.9 & 87.2 \\
SDSTrack~\citep{hou2024sdstrack} & CVPR24 & 53.1 & 66.5 & 62.5 & 84.8 & 59.7 & 76.7 & 72.8 & 81.2 & 88.3 \\
UnTrack~\citep{wu2024untrack} & CVPR24 & 51.3 & 64.6 & 62.5 & 84.2 & 58.9 & 75.5 & 71.8 & 82.0 & 86.4 \\
CMDTrack~\citep{zhang2025cross} & TPAMI25 & 56.6 & 68.8 & 61.8 & 85.9 & 61.3 & 75.8 & -- & -- & -- \\
STTrack~\citep{hu2025sttrack} & AAAI25 & 60.3 & 76.0 & 66.7 & 89.8 & 61.9 & 78.6 & \textbf{77.6} & 82.5 & \textbf{93.7} \\
SMSTracker~\citep{chan2025smstracker} & ICCV25 & 56.0 & 70.3 & 64.5 & 86.9 & 60.4 & 76.3 & 74.8 & 82.2 & 89.7 \\
XTrack~\citep{tan2025xtrack} & ICCV25 & 55.7 & 69.1 & 64.9 & 87.4 & 60.9 & 77.5 & 74.0 & 82.8 & 88.9 \\
SUTrack~\citep{chen2025sutrack} & AAAI25 & 59.9 & 74.5 & 69.5 & 92.2 & 62.7 & 79.9 & 76.5 & 82.8 & 91.8 \\
UniSOT~\citep{ma2026unisot} & TPAMI26 & 54.0 & 67.4 & 64.1 & 86.6 & 60.7 & 78.0 & 73.2 & 81.8 & 88.8 \\
SEATrack~\citep{su2026seatrack} & CVPR26 & 57.3 & 71.6 & 63.9 & 87.8 & 60.3 & 77.1 & 73.6 & 82.1 & 88.4 \\
UETrack~\citep{Dong_2026_CVPR} & CVPR26 & 55.5 & 69.1 & 64.2 & 86.7 & 59.2 & 76.2 & 68.3 & 80.8 & -- \\
UTPTrack~\citep{Wu_2026_CVPR} & CVPR26 & 59.2 & 73.9 & \underline{70.4} & 92.9 & 62.6 & 80.0 & 76.4 & 82.7 & 92.1 \\
TGTrack~\citep{geng2026tgtrack} & CVPR26 & \underline{61.7} & \textbf{77.4} & 69.9 & \underline{93.7} & \underline{63.9} & \underline{81.4} & \underline{77.4} & 82.9 & \underline{93.0} \\
OneTrackerV2~\citep{hong2026onetrackerv2} & ICML26 & 60.6 & 75.5 & 69.5 & 91.3 & 63.0 & 79.8 & -- & -- & -- \\
\midrule
UniMergeTrack & Ours & \textbf{61.8} & \underline{77.2} & \textbf{70.6} & \textbf{94.0} & \textbf{64.1} & \textbf{81.5} & 76.8 & \textbf{83.1} & 92.3  \\
\bottomrule
\end{tabular}
}
\end{table*}

\begin{table*}[t]
\centering
\caption{Generalization comparison on COESOT and CDTB datasets.}
\label{tab:generalization}
\renewcommand{\arraystretch}{1.08}
\setlength{\tabcolsep}{6pt}
\resizebox{\textwidth}{!}{
\begin{tabular}{c|c|ccccccc}
\toprule
Dataset & Metric  & SMSTracker & XTrack & SUTrack & SEATrack & UETrack & UTPTrack & UniMergeTrack \\
\hline
\multirow{2}{*}{\shortstack{COESOT\\(RGB-Event)}}
& SR  & 63.6 & \underline{65.2} & 64.6 & 64.8 & 61.5 & 64.7 & \textbf{66.3} \\
& PR  & 76.8 & 78.8 & 77.7 & \underline{78.9} & 74.7 & 78.2 & \textbf{80.5} \\
\hline
\multirow{2}{*}{\shortstack{CDTB\\(RGB-Depth)}}
& F-score   &69.7  & 69.8  & 70.8 &68.5  &67.7   &\underline{71.5}  &\textbf{72.0}  \\
& Precision &70.5  & 68.5  & 69.8 &69.4  &68.2   &\underline{71.4}  &\textbf{72.0}  \\
\bottomrule
\end{tabular}
}
\end{table*}

\subsection{Ablation Study}

$\bullet$ \textbf{Component Analysis.~}
As shown in Table~\ref{tab:component_analysis}, we conduct a component analysis on both LasHeR and VisEvent to evaluate the contribution of each module.
Specifically, when only the first-stage task-specific training is performed without model merging, the average AUC and PR decrease by 2.5 and 3.1 percentage points, respectively, demonstrating the benefit of integrating knowledge across RGB-X tasks.
In addition, removing distribution alignment fine-tuning causes significant performance degradation, highlighting the importance of alleviating the distribution shift induced by model merging to ensure effective collaboration between the unified backbone and tracking heads.

For the hierarchical merging design, removing high-pass filtering, the lightweight modality expert, or intra-task tracking head merging results in average PR drops of 1.8, 1.5, and 1.4 percentage points, respectively, validating their crucial roles in preserving transferable and task-specific knowledge.
Finally, removing optimal transport-guided channel alignment reduces average PR by 0.9 percentage points, demonstrating its benefit for reliable intra-task tracking head merging.


\begin{table}[t]
\centering
\small
\caption{Component Analysis of UniMergeTrack on LasHeR and VisEvent. $\Delta$ denotes the average performance change relative to the full model.}
\label{tab:component_analysis}
\setlength{\tabcolsep}{4pt}
\resizebox{0.95\columnwidth}{!}{
\begin{tabular}{c|l|cccc|cc}
\hline
\multirow{2}{*}{\textbf{No.}} & \multirow{2}{*}{\textbf{Variant}} & \multicolumn{2}{c}{\textbf{LasHeR}} & \multicolumn{2}{c|}{\textbf{VisEvent}} & \multicolumn{2}{c}{\textbf{Avg. Change}} \\
\cline{3-8}
& & \textbf{AUC} & \textbf{PR} & \textbf{AUC} & \textbf{PR} & $\boldsymbol{\Delta}$\textbf{AUC} & $\boldsymbol{\Delta}$\textbf{PR} \\
\hline
1 & \textbf{UniMergeTrack (Ours)} & \textbf{61.8} & \textbf{77.2} & \textbf{64.1} & \textbf{81.5} & \textbf{--} & \textbf{--} \\
2 & w/o Model Merging (Baseline) & 58.8 & 73.4 & 62.2 & 79.2 & -2.5 & -3.1 \\
3 & w/o Distribution Alignment Fine-Tuning & 57.7 & 73.2 & 60.9 & 78.6 & -3.7 & -3.5 \\
\hline
4 & w/o High-pass Filtering & 60.1 & 75.2 & 62.9 & 80.0 & -1.5 & -1.8 \\
5 & w/o Lightweight Modality Expert & 60.3 & 75.3 & 63.0 & 80.4 & -1.3 & -1.5 \\
6 & w/o Intra-Task Tracking Head Merging & 60.4 & 75.6 & 63.2 & 80.3 & -1.2 & -1.4 \\
7 & w/o Optimal Transport Channel Alignment & 61.0 & 76.2 & 63.5 & 80.8 & -0.7 & -0.9 \\

\hline
\end{tabular}
}
\end{table}

$\bullet$ \textbf{Analysis of Different Training Paradigms.~} 
As shown in Table~\ref{tab:paradigm_comparison}, joint training consistently outperforms independent training, confirming the benefit of cross-task knowledge sharing. Compared with joint training, the model merging paradigm further improves PR by 2.7 and 1.6 points on LasHeR and VisEvent, respectively, achieving the best overall performance.

$\bullet$ \textbf{Analysis of Different Merging Strategies.~}
As shown in Table~\ref{tab:merging_strategies}, we compare our hierarchical merging strategy with representative model merging methods. Our method achieves the best performance on both datasets, outperforming FREE-Merging by 1.6 and 1.2 PR on LasHeR and VisEvent, respectively. These results validate the effectiveness of our hierarchical merging strategy.

\begin{table*}[t]
\centering

\begin{minipage}[t]{0.45\textwidth}
\centering
\scriptsize
\captionof{table}{Analysis of Different Training Paradigms on LasHeR and VisEvent.}
\label{tab:paradigm_comparison}
\setlength{\tabcolsep}{4pt}
\renewcommand{\arraystretch}{0.95}
\resizebox{\linewidth}{!}{
\begin{tabular}{l|cccc}
\hline
\multirow{2}{*}{\textbf{Paradigm}} 
& \multicolumn{2}{c}{\textbf{LasHeR}} 
& \multicolumn{2}{c}{\textbf{VisEvent}} \\
\cline{2-5}
& \textbf{AUC} & \textbf{PR} & \textbf{AUC} & \textbf{PR} \\
\hline
Independent Training & 58.8 & 73.4 & 62.2 & 79.2 \\
Joint Training       & 59.9 & 74.5 & 62.7 & 79.9 \\
\textbf{Model Merging (Ours)}  & \textbf{61.8} & \textbf{77.2} & \textbf{64.1} & \textbf{81.5} \\
\hline
\end{tabular}
}
\end{minipage}
\hfill
\begin{minipage}[t]{0.52\textwidth}
\centering
\scriptsize
\captionof{table}{Analysis of Different Merging Strategies on LasHeR and VisEvent.}
\label{tab:merging_strategies}
\setlength{\tabcolsep}{3.5pt}
\renewcommand{\arraystretch}{0.95}
\resizebox{\linewidth}{!}{
\begin{tabular}{l|cccc}
\hline
\multirow{2}{*}{\textbf{Merging Strategy}}
& \multicolumn{2}{c}{\textbf{LasHeR}}
& \multicolumn{2}{c}{\textbf{VisEvent}} \\
\cline{2-5}
& \textbf{AUC} & \textbf{PR} & \textbf{AUC} & \textbf{PR} \\
\hline
Task Arithmetic~\citep{ilharco2023editing}       & 57.6 & 72.4  & 60.6 & 77.7 \\
TIES-Merging~\citep{yadav2023ties}              & 59.2  & 73.6 & 62.5 & 79.2 \\
FREE-Merging~\citep{zheng2025free}               & 60.4 & 75.6 & 63.2 & 80.3 \\
\textbf{Hierarchical Merging (Ours)}             & \textbf{61.8} & \textbf{77.2} & \textbf{64.1} & \textbf{81.5} \\
\hline
\end{tabular}
}
\end{minipage}
\end{table*}

\subsection{Visualization} 
\begin{figure*}[!htp]
\center
\includegraphics[width=\linewidth]{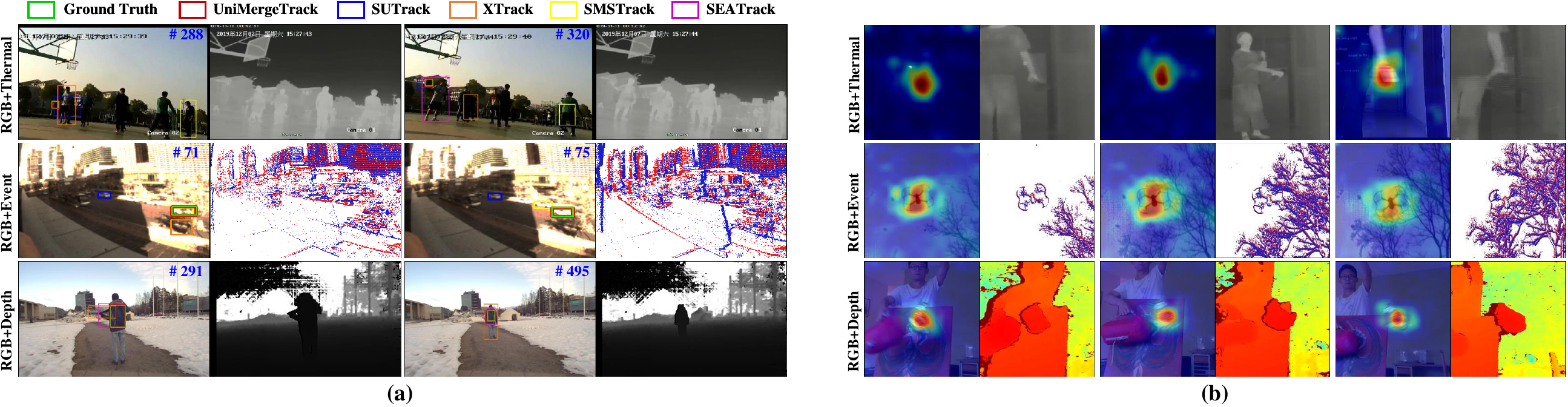}
\caption{Visualization of (a) tracking results and (b) attention maps produced by UniMergeTrack.} 
\label{Tracking_Results_Attn_maps}
\end{figure*}

In this section, we present qualitative visualizations to provide a more intuitive understanding of our method. Figure~\ref{Tracking_Results_Attn_maps} (a) presents comparisons of tracking results across RGB-Thermal, RGB-Event, and RGB-Depth tracking tasks. UniMergeTrack exhibits more accurate and stable target localization across these diverse multimodal tracking scenarios. We further visualize the attention maps in Figure~\ref{Tracking_Results_Attn_maps} (b), which mainly focus on target regions while suppressing irrelevant backgrounds. These visualizations suggest that the proposed model achieves robust target localization in challenging scenarios while effectively capturing target-relevant information.

\section{Conclusion} 
In this work, we propose UniMergeTrack, the first model merging paradigm for RGB-X tracking. UniMergeTrack combines task-specific optimization with cross-task knowledge integration through hierarchical model merging, where backbones are merged across tasks and tracking heads are merged within each task. A subsequent distribution alignment stage further establishes compatibility between the unified backbone and task-specific heads. Extensive experiments across RGB-Thermal, RGB-Event, and RGB-Depth tracking demonstrate superior performance and strong generalization to unseen data. Our results highlight model merging as a promising paradigm for building unified RGB-X trackers while mitigating knowledge isolation and optimization interference.


\section*{LLM Usage}
We used LLMs to improve the grammar, phrasing, and readability of this manuscript. We also used LLMs as a search tool to help us find related works, but we independently reviewed, interpreted, and summarized all identified works.

\bibliography{iclr2027_conference}
\bibliographystyle{iclr2027_conference}

\appendix

\section*{\textbf{Appendix}}

\section{Gradient Conflict Rate Calculation}
Multi-task learning facilitates knowledge sharing across tasks, but joint optimization can also introduce interference due to competing task objectives. To quantify such interference caused by joint training in RGB-X tracking, we measure pairwise gradient conflicts using the Gradient Conflict Rate (GCR), defined based on the cosine similarity between task-specific gradients~\citep{yu2020pcgrad,gong2022nasvit}.

Given a task $i$, we compute its gradient with respect to the trainable parameters $\theta$ of the shared encoder as
\begin{equation}
    g_i = \nabla_{\theta}\mathcal{L}_i,
\end{equation}
where $\mathcal{L}_i$ denotes the training loss for task $i$. For a pair of tasks $i$ and $j$, the cosine similarity between their gradients is defined as
\begin{equation}
    C_{ij}
    =
    \frac{g_i^{\top}g_j}
    {\|g_i\|_2\|g_j\|_2}.
\end{equation}
We consider the two tasks to exhibit a gradient conflict when $C_{ij}<0$, indicating opposing optimization directions on the shared encoder.

We evaluate gradient conflicts every 20 epochs throughout joint training. At each evaluation checkpoint, we construct $M=1000$ batch triplets, each consisting of independently sampled batches from the RGB-T, RGB-D, and RGB-E training sets. For the $m$-th triplet, we separately compute the encoder gradients induced by the three tasks using the same model checkpoint, without updating the model parameters. We then obtain the following three pairwise cosine similarities:
\begin{equation}
    C_{TD}^{(m)}, \qquad
    C_{TE}^{(m)}, \qquad
    C_{DE}^{(m)}.
\end{equation}

For each task pair $(i,j)$, the Gradient Conflict Rate is defined as the proportion of sampled batch pairs that yield a negative gradient cosine similarity:
\begin{equation}
    \mathrm{GCR}_{ij}
    =
    \frac{1}{M}
    \sum_{m=1}^{M}
    \mathbb{I}
    \left(C_{ij}^{(m)} < 0\right)
    \times 100\%,
\end{equation}
where $\mathbb{I}(\cdot)$ denotes the indicator function and $M=1000$ in our experiments. We report the GCR separately for the RGBT-RGBD, RGBT-RGBE, and RGBD-RGBE task pairs. Their average is computed as
\begin{equation}
    \mathrm{GCR}_{\mathrm{mean}}
    =
    \frac{1}{3}
    \left(
    \mathrm{GCR}_{TD}
    +
    \mathrm{GCR}_{TE}
    +
    \mathrm{GCR}_{DE}
    \right).
\end{equation}

A higher GCR indicates more frequent conflicts between task-specific optimization directions on the shared encoder, suggesting stronger cross-task interference during joint training.

\section{Related Work}

\subsection{Optimal Transport}
Optimal transport (OT) provides a principled framework for comparing probability distributions through minimum-cost transport. Kantorovich~\citep{kantorovich1942translocation} relaxed the original formulation into a tractable linear program, laying the foundation of modern OT, while Benamou and Brenier~\citep{benamou2000computational} introduced a dynamic formulation that interprets transport as continuous mass flow.
A major computational advance came from entropy-regularized OT~\citep{cuturi2013sinkhorn}, which enables efficient optimization via the Sinkhorn algorithm~\citep{sinkhorn1967concerning}. Subsequent work improved scalability through stochastic optimization~\citep{genevay2016stochastic} and extended OT to structured spaces via Gromov-Wasserstein distances~\citep{memoli2011gromov}. More recent studies further improved robustness and scalability through partial and unbalanced formulations~\citep{agarwal2025robust,sun2025variational}, as well as stochastic and structure-aware transport~\citep{nietert2025stochastic,im2026shape}. These advances have established OT as a versatile tool for distribution alignment in modern machine learning~\citep{peyre2019computational}.
During the training of trackers, the channels of the tracking head may drift in parameter space, leading to inconsistent channel correspondences across historical checkpoints. We resort to OT for channel alignment before merging, ensuring more reliable parameter merging.

\section{Implementation Details of OT-Based Tracking Head Merging}

For each task, we use the tracking heads saved at epochs 24 and 30 as the historical head $\boldsymbol{h}_{k}^{\mathrm{his}}$ and the reference head $\boldsymbol{h}_{k}^{\mathrm{ref}}$, respectively. The reference head defines the target channel basis. OT alignment is applied independently to the three branches of the center-based head:
\begin{equation}
\mathcal{B} = \{\mathrm{ctr},\mathrm{offset},\mathrm{size}\},
\end{equation}
where each branch contains five convolutional layers. The following operations are applied independently to each branch.

For layer $l>1$, the historical input channels are first expressed in the reference basis using the transport plan obtained from the preceding layer:
\begin{equation}
\overline{\mathbf{W}}_{l}^{\mathrm{his}}
=
\mathbf{W}_{l}^{\mathrm{his}}
\times_{\mathrm{in}}
\overline{\mathbf{P}}_{l-1},
\end{equation}
where $\mathbf{W}_{l}^{\mathrm{his}}$ and $\overline{\mathbf{W}}_{l}^{\mathrm{his}}$ are the historical weights before and after input-channel alignment, respectively, and $\overline{\mathbf{P}}_{l-1}$ is the normalized transport plan. For the first layer, we set
\begin{equation}
\overline{\mathbf{W}}_{1}^{\mathrm{his}} = \mathbf{W}_{1}^{\mathrm{his}}.
\end{equation}

Each output channel is represented by its normalized incoming weights, augmented with the standardized convolutional bias and BatchNorm statistics when available:
\begin{equation}
\mathbf{d}_{l,i}^{\mathrm{his}}
=
\operatorname{Norm}_2
\left(
\left[
\operatorname{Norm}_2
\left(
\operatorname{vec}
(\overline{\mathbf{W}}_{l,i}^{\mathrm{his}})
\right);
\beta\widehat{\mathbf{s}}_{l,i}^{\mathrm{his}}
\right]
\right),
\end{equation}

\begin{equation}
\mathbf{d}_{l,j}^{\mathrm{ref}}
=
\operatorname{Norm}_2
\left(
\left[
\operatorname{Norm}_2
\left(
\operatorname{vec}
(\mathbf{W}_{l,j}^{\mathrm{ref}})
\right);
\beta\widehat{\mathbf{s}}_{l,j}^{\mathrm{ref}}
\right]
\right).
\end{equation}
where $\mathbf{W}_{l,i}$ is the kernel of the $i$-th output channel in layer $l$, $\operatorname{vec}(\cdot)$ flattens the kernel, and $\operatorname{Norm}_2(\cdot)$ denotes $\ell_2$ normalization. The vector $\widehat{\mathbf{s}}_{l,i}$ contains the corresponding standardized bias and BatchNorm statistics, while $[\cdot\,;\cdot]$ denotes vector concatenation. The coefficient $\beta$ controls the contribution of these auxiliary statistics, and we set $\beta=0.1$.

The pairwise alignment cost between the historical channel $i$ and the reference channel $j$ is defined as
\begin{equation}
C_{l}^{ij} =
\left\|
\mathbf{d}_{l,i}^{\mathrm{his}}
-
\mathbf{d}_{l,j}^{\mathrm{ref}}
\right\|_2^2.
\label{eq:head_alignment_cost}
\end{equation}

We obtain the channel correspondence by solving the entropy-regularized optimal transport problem:
\begin{equation}
\mathbf{P}_{l}^{*}
=
\arg\min_{\mathbf{P}\geq 0}
\left\langle
\mathbf{P},\mathbf{C}_{l}
\right\rangle_F
-
\epsilon\mathcal{H}(\mathbf{P}),
\end{equation}
subject to
\begin{equation}
\mathbf{P}\mathbf{1}
=
\frac{1}{B_s}\mathbf{1},
\qquad
\mathbf{P}^{\top}\mathbf{1}
=
\frac{1}{B_t}\mathbf{1},
\label{eq:head_ot}
\end{equation}
where $B_s$ and $B_t$ denote the numbers of historical and reference channels, respectively. We use Sinkhorn iterations~\citep{cuturi2013sinkhorn} with $\epsilon=0.05$, at most 300 iterations, and a stopping tolerance of $10^{-6}$. Because the two heads originate from the same task and training trajectory, an identity prior of 0.02 is added to the off-diagonal transport costs to discourage unnecessary channel permutations.

After column normalization, we denote the normalized transport plan by $\overline{\mathbf{P}}_{l}$, which maps the historical output channels to the reference basis:
\begin{equation}
\widetilde{\mathbf{W}}_{l}^{\mathrm{his}}
=
\overline{\mathbf{P}}_{l}^{\top}
\overline{\mathbf{W}}_{l}^{\mathrm{his}}.
\label{eq:head_output_alignment}
\end{equation}
The same mapping is applied to the corresponding bias and BatchNorm parameters. The resulting transport plan $\overline{\mathbf{P}}_{l}$ is then propagated to
the input channels of layer $l+1$.

OT-based output-channel alignment is applied to the first four layers. For the fifth layer, only the input channels are aligned because its output channels have fixed semantics: objectness for the \texttt{ctr} branch, $(\Delta x,\Delta y)$ for the \texttt{offset} branch, and $(w,h)$ for the \texttt{size} branch.

After alignment, the historical and reference heads are merged by weighted averaging:
\begin{equation}
\boldsymbol{H}_{k}
=
\alpha
\boldsymbol{h}_{k}^{\mathrm{ref}}
+
(1-\alpha)
\widetilde{\boldsymbol{h}}_{k}^{\mathrm{his}}.
\label{eq:head_weighted_merging}
\end{equation}
We set $\alpha$ to $0.5$, $0.7$, and $0.5$ for RGB-Thermal, RGB-Event, and RGB-Depth tracking, respectively.

\section{More Experiments and Analysis}

\subsection{Datasets and Evaluation Metric}

\textbf{RGB-Thermal Tracking.}
For RGB-Thermal tracking, we adopt LasHeR and RGBT234. LasHeR is a large-scale RGB-Thermal tracking benchmark containing 1,224 aligned RGB-Thermal video pairs with diverse object categories and environmental conditions. RGBT234 consists of 234 RGB-Thermal sequences, which is widely used for evaluating the robustness of RGBT trackers. Following the standard evaluation protocols, we report AUC (Area Under the Success Curve), PR (Precision Rate), MSR (Mean Success Rate), and MPR (Mean Precision Rate). Specifically, AUC measures the area under the success plot and reflects the overall tracking performance across different overlap thresholds. PR evaluates localization accuracy by measuring the proportion of frames with accurate target center predictions. MSR and MPR denote the average success rate and precision rate over all sequences. We train UniMergeTrack on LasHeR and evaluate it on both LasHeR and RGBT234.

\textbf{RGB-Event Tracking.}
For RGB-Event tracking, we use VisEvent and COESOT. VisEvent contains 820 RGB-Event video pairs collected under challenging conditions, such as low illumination, fast motion, and background clutter. COESOT is a large-scale RGB-Event tracking benchmark containing diverse scenarios. We follow the standard protocols and report AUC and PR for evaluation. UniMergeTrack is trained on VisEvent and evaluated on both VisEvent and COESOT to assess its generalization ability across RGB-Event datasets.

\textbf{RGB-Depth Tracking.}
For RGB-Depth tracking, we adopt DepthTrack, VOT-RGBD22, and CDTB. DepthTrack is a large-scale RGB-Depth tracking benchmark used for training RGB-Depth trackers, while VOT-RGBD22 and CDTB provide challenging evaluation scenarios with diverse RGB-Depth sequences. VOT-RGBD22 follows the VOT evaluation protocol and reports EAO (Expected Average Overlap), Accuracy (Acc.), and Robustness (Rob.), which measure overall tracking performance, localization accuracy, and tracking stability, respectively. CDTB evaluates RGB-Depth trackers using Precision and F-score, where Precision reflects localization accuracy and F-score provides a balanced evaluation by combining Precision and Recall. We train UniMergeTrack on DepthTrack and evaluate it on VOT-RGBD22 and CDTB. We do not report results on the DepthTrack test set because its evaluation toolkit has been updated and the original version is no longer available, preventing the reproduction of existing methods and making fair comparisons infeasible.

\subsection{Supplementary Experiments}

$\bullet$ \textbf{Analysis of Hierarchical Merging Strategy.~} 
As shown in Table~\ref{tab:hierarchical_ablation}, cross-task backbone merging significantly improves the baseline by transferring shared knowledge across tasks. In contrast, intra-task head merging outperforms cross-task head merging, achieving the best results on both datasets. These results support our design of cross-task backbone merging and intra-task head merging.

$\bullet$ \textbf{Ablation Study of Different Channel Alignment Strategies.~} 
Table~\ref{tab:alignment_ablation} compares different channel alignment strategies for intra-task tracking head merging. Direct parameter averaging suffers from hidden-channel inconsistency, while Hungarian alignment is limited by hard matching. OT-based alignment achieves the best performance on both datasets.

\begin{table}[t]
\centering
\small

\begin{minipage}{0.49\columnwidth}
\centering
\caption{Analysis of Hierarchical Merging Strategy on LasHeR and VisEvent.}
\label{tab:hierarchical_ablation}
\setlength{\tabcolsep}{3pt}
\resizebox{\columnwidth}{!}{
\begin{tabular}{c|cc|cccc}
\hline
\multirow{2}{*}{\textbf{No.}} &
\multicolumn{2}{c|}{\textbf{Merging Strategy}} &
\multicolumn{2}{c}{\textbf{LasHeR}} &
\multicolumn{2}{c}{\textbf{VisEvent}} \\
\cline{2-7}
&
\textbf{Backbone} &\textbf{Tracking Head} &
\textbf{AUC} &\textbf{PR} &
\textbf{AUC} &\textbf{PR} \\
\hline
1 & -- & -- & 58.8 & 73.4 & 62.2 & 79.2 \\
2 & Cross-task & -- & 60.4 & 75.6 & 63.2 & 80.3 \\
3 & Cross-task & Cross-task & 60.7 & 76.0 & 63.4 & 80.6 \\
4 & \textbf{Cross-task} & \textbf{Intra-task} & \textbf{61.8} & \textbf{77.2} & \textbf{64.1} & \textbf{81.5} \\
\hline
\end{tabular}
}
\end{minipage}
\hfill
\begin{minipage}{0.47\columnwidth}
\centering
\caption{Ablation Study of Different Channel Alignment Strategies.}
\label{tab:alignment_ablation}
\setlength{\tabcolsep}{3pt}
\resizebox{\columnwidth}{!}{
\begin{tabular}{l|cc|cc}
\hline
\multirow{2}{*}{\textbf{Merging Strategy}} 
& \multicolumn{2}{c|}{\textbf{LasHeR}} 
& \multicolumn{2}{c}{\textbf{VisEvent}} \\
\cline{2-5}
& \textbf{AUC} & \textbf{PR} 
& \textbf{AUC} & \textbf{PR} \\
\hline
w/o Alignment & 61.0 & 76.2 & 63.5 & 80.8 \\
Hungarian Alignment & 61.2 & 76.5 & 63.5 & 81.0 \\
\textbf{OT Alignment (Ours)} & \textbf{61.8} & \textbf{77.2} & \textbf{64.1} & \textbf{81.5} \\
\hline
\end{tabular}
}
\end{minipage}

\end{table}

\begin{figure*}[!htp]
\center
\includegraphics[width=\linewidth]{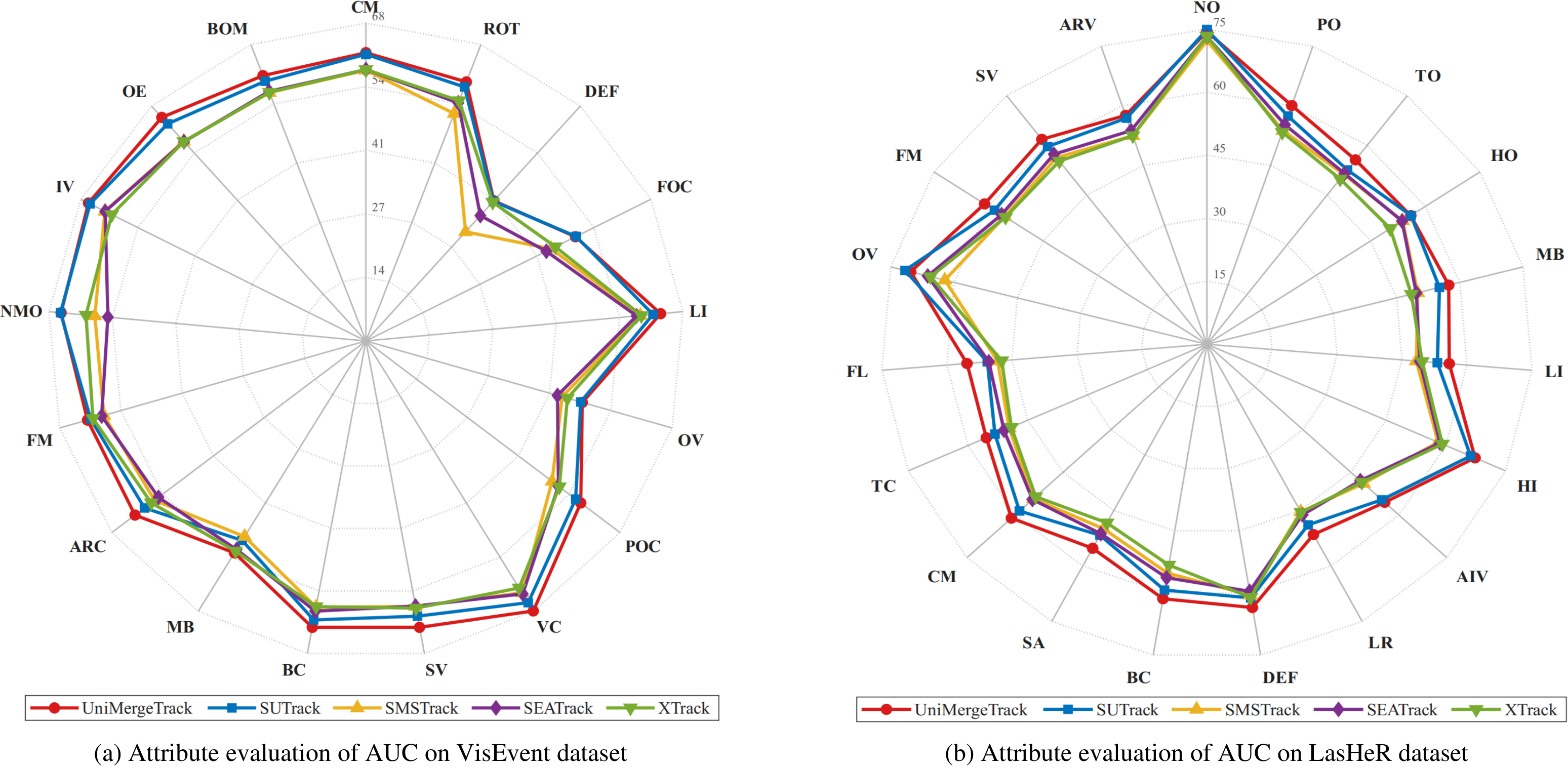}
\caption{Attribute evaluation on the VisEvent and LasHeR test sets.} 
\label{fig:attribute}
\end{figure*}

$\bullet$ \textbf{Attribute Evaluation on the VisEvent and LasHeR Datasets.}
As shown in Fig.~\ref{fig:attribute}, we present the attribute evaluation on VisEvent and LasHeR. UniMergeTrack achieves consistently competitive performance across diverse challenging attributes, including Scale Variation, Viewpoint Change, Low Illumination, and Background Clutter. These improvements demonstrate that our model merging strategy preserves task-specific knowledge while promoting cross-task integration, enabling robust generalization across diverse tracking conditions.

\subsection{Hyperparameter Settings}
As shown in Table~\ref{tab:hyper_parameters}, we use task-specific hyperparameters for different RGB-X tracking scenarios. The expert sparsity $d$ is set to 0.01/0.02/0.02, and the reference head contribution coefficient $\alpha$ is set to 0.5/0.7/0.5 for RGB-Thermal, RGB-Event, and RGB-Depth tracking, respectively.

\begin{table}[t]
\centering
\caption{Hyperparameter settings for different tasks.}
\label{tab:hyper_parameters}
\begin{tabular}{c|ccc}
\hline
\textbf{Hyperparameter} & \textbf{RGB-Thermal} & \textbf{RGB-Event} & \textbf{RGB-Depth} \\
\hline
Expert sparsity ($d$) & 0.01 & 0.02 & 0.02 \\
Reference head contribution ($\alpha$) & 0.50 & 0.70 & 0.50 \\
\hline
\end{tabular}
\end{table}

\subsection{Efficiency Analysis} 
UniMergeTrack operates entirely in the parameter space without altering the network architecture, thereby preserving the original model complexity and inference efficiency at 89M parameters, 23 GFLOPs, and 55 FPS. In terms of storage, it shares the merged backbone across tasks while retaining lightweight task-specific components, reducing the storage requirement from \(3 \times 340\) MB for independently trained models to 379 MB, only 39 MB more than joint training (340 MB).

\section{Limitation Analysis} 
Although UniMergeTrack achieves effective multimodal tracking, there also remains room for further improvement. First, the current framework does not incorporate the text modality. Text-guided multimodal tracking, potentially integrated with multimodal large language models, represents a promising direction for enhancing the model’s target perception. Second, the framework does not explicitly account for missing or degraded modalities in challenging scenarios. Developing quality-aware and adaptive multimodal fusion mechanisms could further improve tracking robustness to incomplete or unreliable multimodal inputs.

\end{document}

%% file: math_commands.tex
\usepackage{amsmath,amsfonts,bm}

\def\eqref#1{equation~\ref{#1}}

\def\1{\bm{1}}

\DeclareMathAlphabet{\mathsfit}{\encodingdefault}{\sfdefault}{m}{sl}
\SetMathAlphabet{\mathsfit}{bold}{\encodingdefault}{\sfdefault}{bx}{n}

